\documentclass[conference]{IEEEtran}
\usepackage{amsmath,amssymb,amsfonts}
\usepackage{graphicx}
\usepackage{textcomp}
\usepackage[ruled,vlined]{algorithm2e}
\usepackage{amsmath}
\usepackage{booktabs}
\usepackage{array}
\usepackage{multirow}
\usepackage{newfloat}
\usepackage{url}
\usepackage[dvipsnames]{xcolor}
\usepackage[numbers]{natbib}
\begin{document}

\definecolor{lightgray}{rgb}{0.9, 0.9, 0.9}
\definecolor{ballblue}{rgb}{0.13, 0.67, 0.8}
\definecolor{lightgreen}{rgb}{0.95, 0.95, 0.95}
\definecolor{gg}{HTML}{e2f0cb}

\newcommand{\STAB}[1]{\begin{tabular}{@{}c@{}}#1\end{tabular}}
\newcommand{\tabstyle}[1]{
  \setlength{\tabcolsep}{#1}
  \renewcommand{\arraystretch}{\tableCellHeight}
  \centering
  \small
}

\title{Pear: Pruning and Sharing Adapters in Visual Parameter-Efficient Fine-Tuning}

\author{\IEEEauthorblockN{1\textsuperscript{st} Yibo Zhong}
\textit{Computer Science} \\
\IEEEauthorblockA{\textit{Sichuan University} \\
Chengdu, China \\
zhongyibo@stu.scu.edu.cn}
\and
\IEEEauthorblockN{2\textsuperscript{nd} Yao Zhou*}
\textit{Computer Science} \\
\IEEEauthorblockA{\textit{Sichuan University} \\
Chengdu, China \\
yaozhou@scu.edu.cn}
}

\maketitle

\begin{abstract}
Adapters have been widely explored to alleviate computational and storage costs when fine-tuning pretrained foundation models. However, the adapter itself can exhibit redundancy, leading to unnecessary storage overhead and inferior performance. In this paper, we propose \textbf{P}run\textbf{e} and Sh\textbf{ar}e (Pear), a novel adapter-pruning framework for efficient fine-tuning of pretrained visual foundation models. Specifically, we prune certain adapters and share the more important unpruned ones with positions where adapters are pruned, allowing continual adaptation at these positions after pruning. Additionally, a knowledge checkpoint strategy is introduced, which preserves the information of the pruned adapters and further boosts performance. Experimental results on visual adaptation benchmark validate the effectiveness and efficiency of the proposed Pear comparing to other competitive methods. Code is in \url{https://github.com/yibozhong/pear}.
\end{abstract}


\section{Introduction}

Foundation models like Transformer have found widespread application across various fields. A typical application involves fine-tuning pretrained models on downstream tasks specific to a particular domain. However, when fine-tuning the entire model, known as \textit{Full Fine-Tuning} (FFT), the process can be slow or even infeasible due to the large number of parameters in the model and the constraints of computational resources. The required high GPU memory for FFT often poses a significant challenge. To address this issue, Parameter-Efficient Fine-Tuning (PEFT) has been proposed. The core of PEFT lies in introducing lightweight learnable prompts \citep{jia2022VPT} \citep{lu2022promptDistributionLearning} \citep{zhou2022learningToPromptForVision-LanguageModeels_prompt_based_method5} \citep{shen2024multitask_prompt_based_method3} \citep{zang2022unified_prompt_based_method4} \citep{zhou2022conditional_prompt_based_method_6} or adapters modules. The latter has become a major research focus in PEFT. Many adapter-based PEFT methods have been widely adopted \citep{hu2021lora} \citep{chen2022adaptformer} \citep{houlsby2019parameter} \citep{jie2023fact} \citep{jie2022convolutional} \citep{pfeiffer2020adapter-P}, significantly alleviating the computational burden and GPU memory occupancy limitations previously associated with FFT.

\begin{figure}[t]
    \centering
    \includegraphics[width=0.95\linewidth]{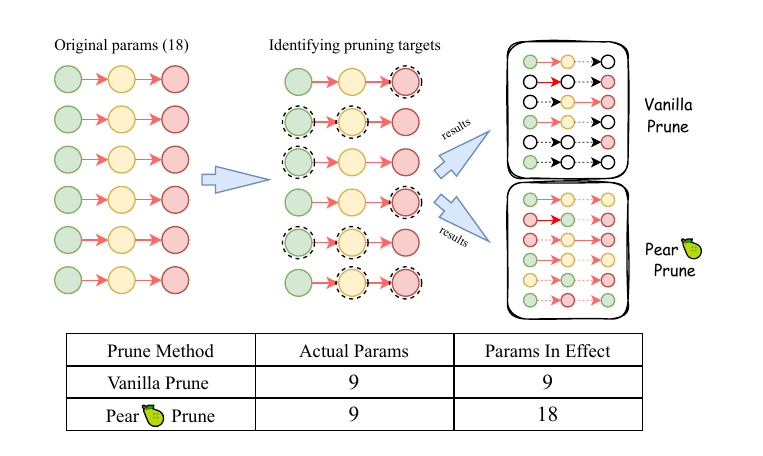}
    \caption{Comparison between vanilla structural pruning techniques and Pear pruning. For the example provided, we begin with a total of 18 parameters before pruning and aim to prune 50\% of them. We first analyze the pruning process of these two methods, which are essentially the same in the first two phases but diverge in the third. Instead of simply pruning the redundant adapters and leaving the corresponding positions unadapted by adapters, Pear shares the unpruned adapters with these positions. Since we use one color to indicate the level of a adapter, the red and yellow adapters in the Pear result is obtained by sharing unpruned red and yellow adapters from other layers. This allows for further adaptation while avoiding any additional parameters. The term 'actual params' indicates the number of parameters after pruning, while 'params in effect' indicates the number of positions that are being adapted.}
    \label{fig:pear}
\end{figure}

The good performance of PEFT stems from the redundancy of the model's extensive parameters. This redundancy is not only present in the model itself, but also in the adapters used in PEFT \citep{zhang2023adalora}. Consequently, a straightforward idea is to employ structural pruning methods to remove redundant adapters during the PEFT process. This approach can further reduce GPU memory usage and enhance storage efficiency. We refer to this method as vanilla pruning, as shown in Fig. \ref{fig:pear}.

However, vanilla pruning still has its shortcomings: structural pruning directly removes the corresponding structures, resulting in positions where adapters are pruned receiving no adaptation at all. In this paper, we propose \textbf{P}run\textbf{e} and Sh\textbf{ar}e (Pear): a novel pruning method that enables continual adaptation for all positions of the model after structural pruning without adding any extra parameters. Therefore, Pear maintains the same parameter size as vanilla pruning, preserving the advantages of reduced memory requirements and storage efficiency that come with fewer parameters. At the same time, it ensures continual adaptation for all positions, resulting in a more comprehensive adaptation and, consequently, often better performance.

\begin{figure*}[t]
    \centering
    \includegraphics[width=0.95\linewidth]{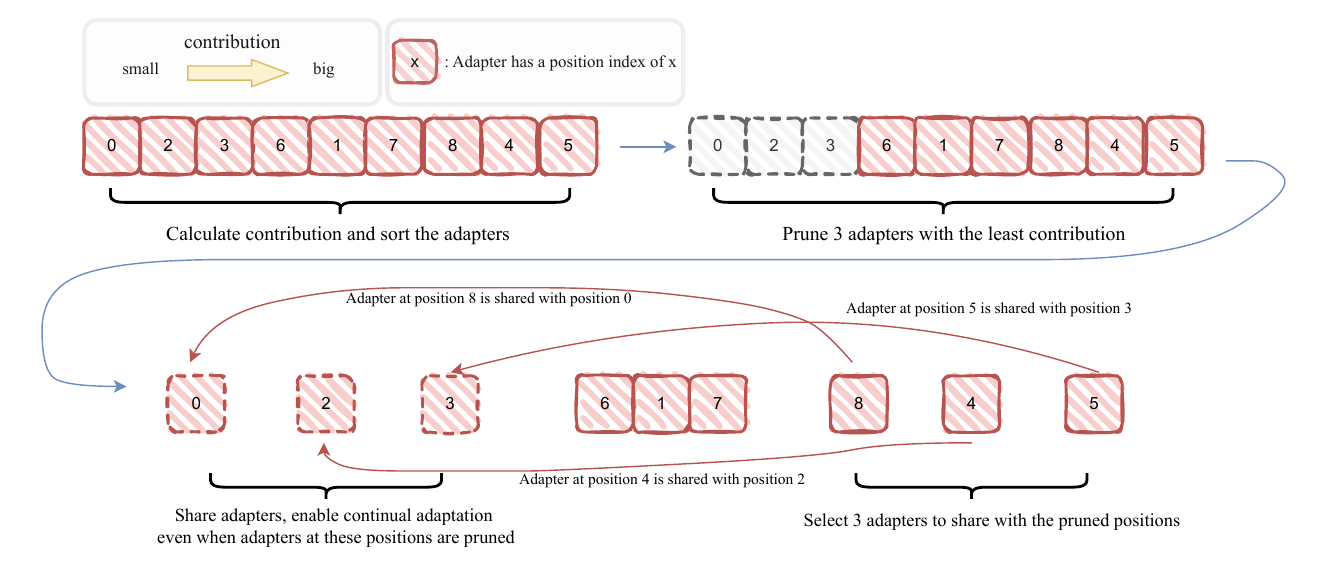}
    \vspace{-5pt}
    \caption{Illustration of the process of \textit{prune and share} of Pear. The two low-rank adapters are combined and considered as a whole. The adapters at each position are first sorted based on their contributions, just like in vanilla pruning. For the example here we aim to prune 3 adapters in total, therefore we choose the 3 adapters with the least contribution (i,e, 3 left most adapters). After pruning, the 3 positions of index 0, 2 and 3 don't have any adaptation in vanilla pruning, which despite their seemingly trivial contribution, still could influence the overall adaptation performance. Pear tackles this issue by share the 3 adapters that are most contribute with these positions to enable continual adaption even when adapters in these positions are pruned. Additional proposed technique by Pear like knowledge checkpoint can also be employed under this framework.}
    \label{fig:process}
\end{figure*}

Since directly pruning adapters also means discarding the information they contain, even though these adapters are relatively unimportant, retaining their information could still be beneficial for comprehensive adaptation. We refer to the retention of information at the positions of pruned adapters as knowledge checkpoint, and we have developed several variants within the Pear framework based on whether and how knowledge checkpoint is utilized, offering more flexible configurations and often superior performance. Our contributions are:
\begin{itemize}
    \item We propose \textbf{P}run\textbf{e} and Sh\textbf{ar}e (Pear), a novel pruning method that supports continuous adaptation at every model position following structural pruning, all while not introducing any supplementary parameters compared to vanilla pruning.
    \item  We propose a strategy termed as knowledge checkpoint, which retains the information of the pruned adapters, thereby benefiting comprehensive and performative adaptation. 
    \item We evaluate Pear on tasks from VTAB-1K benchmark, validating its efficiency and good performance.
\end{itemize}

\section{Methodology}

We first present Pear's two-step pipeline of adaptation, which includes the identification of redundant parameters and then the process of \textit{prune and share} using Pear's framework. Finally, we provide the definition of knowledge checkpoint and discuss its application in Pear.

\subsection{Pruning Pipeline of Pear}

Given a pruning problem, we first define the set of overall parameters $P$ and the target ratio $ratio$ for pruning. Since our paper focuses on the problem of parameter-efficient fine-tuning (PEFT in short), specifically adapter-based PEFT techniques, we define $P$ as all the inserted adapters. In LoRA \citep{hu2021lora}, an adaptation to a weight matrix $W \in \mathbb{R}^{a \times b}$ is performed by inserting two trainable low-rank adapters $A \in \mathbb{R}^{a \times d}$ and $B \in \mathbb{R}^{d \times b}$ where $d$ is a small hidden-dimension such that $d \ll a$ and $d \ll b$, and utilizing their product as an approximation of the actual weight update $\Delta W \in \mathbb{R}^{a \times b}$, which can be expressed as:
\begin{equation}
    W = W_0 + \Delta W = W_0 + AB
\end{equation}
where we denote the value of the frozen weight matrix as $W_0$ and the actual weight during runtime fine-tuning as $W$. In this case, the backbone is kept frozen all the time and $P$ is composed of all the $A$s and $B$s inserted to the frozen model.

\subsubsection{Identifying Redundant Adapters}

To efficiently prune parameters, the ideal approach is to identify parameters that are least important to the current task, also known as redundant parameters. Since the pre-pruning process of Pear is the same as that of vanilla pruning, it can be employed with varying classical pruning criteria used in prior research \citep{kwon2022fast_taylor_prune_for_transformers} \citep{gaikwad2018pruningSqueezenet} \citep{he2018adc_RL_Prune} \citep{luo2017thinetPruningMethodUsingWeight_Greedy1} \citep{molchanov2016pruningForResourseEfficientInference}. 

\subsubsection{Pruning and Sharing}

Once we have identified the redundant parameters, we multiply P size $n$ and the $ratio$ to obtain the total number of parameters $m$ that need to be pruned. Illustrated in Fig. \ref{fig:process}, we select $m$ parameters with the lowest contribution and their related information as the list of parameters to be pruned. To enable sharing, we select $m$ parameters that are not pruned and have the highest contribution for subsequent sharing. Regarding sharing, in the parameters to be pruned, those with a higher contribution will have their positions taken over by the parameters that have a higher contribution among the share list. Utilizing this framework proposed by Pear, we enable \textbf{continual adaptation} to the positions where adapters are pruned \textbf{without any extra parameters incurred}.

\subsection{Knowledge Checkpoint}

Structural pruning often directly removes the original parameters, and information associated with them, such as weights, is also discarded. The consequence of this is that retraining is required to recover performance, and the recovered performance is often sub-optimal. To address this issue, retaining the information of the pruned parameters to aid in adaptation is potentially effective. Under the framework of Pear, we propose a strategy called knowledge checkpoint. Specifically, given a pair of low-rank adapters to be pruned $A$ and $B$, and a pair of adapters to be shared $A\_$ and $B\_$, knowledge checkpoint can be expressed as:
\begin{equation}
    A\_ / B\_  = A\_ / B\_ + A / B
    \label{eq1}
\end{equation}
This straightforward setup directly sum the weight of the pruned adapters to the shared ones. However, as the adapters to be shared have higher contributions, thus their weights should have higher priority, which lead us to a more complex implementation:
\begin{equation}
    A\_ / B\_  = A\_ / B\_ \cdot C_1 + A / B \cdot C_2
    \label{eq2}
\end{equation}
where $C_1$ represents the contribution of the shared adapters while $C_2$ represents the contribution of the pruned adapters. While we could utilize the contribution calculated in the previous step as $C_1$ and $C_2$, we could also manually set them. We term the implementation in Eq. \ref{eq1} as direct aggregation (DA) and the implementation if Eq. \ref{eq2} as score-based aggregation (SBA). 

Utilizing knowledge checkpoint, we can retain the information from the pruned adapters and therefore enable a continual and comprehensive adaptation.

\begin{table*}[t]
\begin{center}
\setlength{\tabcolsep}{0.3pt}
\scalebox{0.90}{
\begin{tabular}{p{3cm}<{}p{1.25cm}<{\centering}p{0.75cm}<{\centering}|p{0.75cm}<{\centering}p{0.75cm}<{\centering}p{0.75cm}<{\centering}p{0.75cm}<{\centering}p{0.75cm}<{\centering}p{0.75cm}<{\centering}p{0.75cm}<{\centering}|p{0.75cm}<{\centering}p{0.75cm}<{\centering}p{0.75cm}<{\centering}p{0.75cm}<{\centering}|p{0.75cm}<{\centering}p{0.75cm}<{\centering}p{0.75cm}<{\centering}p{0.75cm}<{\centering}p{0.75cm}<{\centering}p{0.75cm}<{\centering}p{0.75cm}<{\centering}p{0.75cm}<{\centering}}
\toprule
\multicolumn{3}{c|}{}&\multicolumn{7}{c|}{\textbf{Natural}}&\multicolumn{4}{c|}{\textbf{Specialized}}&\multicolumn{8}{c}{\textbf{Structured}}\\
&\multicolumn{1}{c}{\STAB{\rotatebox[origin=c]{90}{Size (MB)}}}
&\multicolumn{1}{c|}{\STAB{\rotatebox[origin=c]{90}{Avg. Acc.}}}
&\multicolumn{1}{c}{\STAB{\rotatebox[origin=c]{90}{Cifar100}}}
&\multicolumn{1}{c}{\STAB{\rotatebox[origin=c]{90}{Caltech101}}}
&\multicolumn{1}{c}{\STAB{\rotatebox[origin=c]{90}{DTD}}}
&\multicolumn{1}{c}{\STAB{\rotatebox[origin=c]{90}{Flower102}}}
&\multicolumn{1}{c}{\STAB{\rotatebox[origin=c]{90}{Pets}}}
&\multicolumn{1}{c}{\STAB{\rotatebox[origin=c]{90}{SVHN}}}
&\multicolumn{1}{c|}{\STAB{\rotatebox[origin=c]{90}{Sun397}}}
&\multicolumn{1}{c}{\STAB{\rotatebox[origin=c]{90}{Camelyon}}}
&\multicolumn{1}{c}{\STAB{\rotatebox[origin=c]{90}{EuroSAT}}}
&\multicolumn{1}{c}{\STAB{\rotatebox[origin=c]{90}{Resisc45}}}
&\multicolumn{1}{c|}{\STAB{\rotatebox[origin=c]{90}{Retinopathy}}}
&\multicolumn{1}{c}{\STAB{\rotatebox[origin=c]{90}{Clevr-Count}}}
&\multicolumn{1}{c}{\STAB{\rotatebox[origin=c]{90}{Clevr-Dist}}}
&\multicolumn{1}{c}{\STAB{\rotatebox[origin=c]{90}{DMLab}}}
&\multicolumn{1}{c}{\STAB{\rotatebox[origin=c]{90}{KITTI-Dist}}}
&\multicolumn{1}{c}{\STAB{\rotatebox[origin=c]{90}{dSpr-Loc}}}
&\multicolumn{1}{c}{\STAB{\rotatebox[origin=c]{90}{dSpr-Ori}}}
&\multicolumn{1}{c}{\STAB{\rotatebox[origin=c]{90}{sNORB-Azim}}}
&\multicolumn{1}{c}{\STAB{\rotatebox[origin=c]{90}{sNORB-Ele}}}\\
\specialrule{0em}{1pt}{1pt}
\hline
\specialrule{0em}{1pt}{1pt}
\multicolumn{22}{l}{\emph{Conventional Fine-Tuning}}\\
\hline
\specialrule{0em}{1pt}{1pt}
\sc Full&327&68.9&68.9&87.7&64.3&97.2&86.9&87.4&38.8&79.7&95.7&84.2&73.9&56.3&58.6&41.7&65.5&57.5&46.7&25.7&29.1 \\
\sc Linear&0&57.6&64.4&85.0&63.2&97.0&86.3&36.6&51.0&78.5&87.5&68.5&74.0&34.3&30.6&33.2&55.4&12.5&20.0&9.6&19.2\\
\hline
\specialrule{0em}{1pt}{1pt}
\multicolumn{22}{l}{\emph{PEFT methods}}\\
\hline
\specialrule{0em}{1pt}{1pt}
{\sc VPT-Deep}&2.03&72.0&\bf78.8&90.8&65.8&98.0&88.3&78.1&49.6&81.8&\bf96.1&83.4&68.4&68.5&60.0&46.5&72.8&73.6&47.9&32.9&37.8 \\
\text{\color{gray}{\sc NOAH}$^\dag$}&\text{\color{gray}1.37}&\text{\color{gray}75.5}&\text{\color{gray}69.6}&\text{\color{gray}\bf92.7}&\text{\color{gray}70.2}&\text{\color{gray}99.1}&\text{\color{gray}90.4}&\text{\color{gray}86.1}&\text{\color{gray}53.7}&\text{\color{gray}84.4}&\text{\color{gray}95.4}&\text{\color{gray}83.9}&\text{\color{gray}\bf75.8}&\text{\color{gray}82.8}&\text{\color{gray}\bf68.9}&\text{\color{gray}49.9}&\text{\color{gray}\bf81.7}&\text{\color{gray}81.8}&\text{\color{gray}48.3}&\text{\color{gray}32.8}&\text{\color{gray}44.2}\\
{\sc FourierFT}&1.20&72.2&65.3&89.9&69.7&99.0&90.8&81.5&54.8&84.4&93.5&80.6&74.5&73.3&59.0&43.8&77.2&73.7&49.3&26.4&34.0\\
{\sc MoRA}&1.15&75.1&72.6&89.8&72.0&99.1&91.0&89.1&55.7&86.6&94.0&84.7&74.3&78.3&62.8&49.7&78.6&82.2&51.3&34.5&35.1\\
{\sc LoRA}&1.13&76.4&72.0&91.2&71.6&99.1&91.3&88.9&56.4&87.2&94.6&83.9&74.9&83.7&64.0&52.3&81.2&84.8&53.3&\bf38.1&43.4\\
{\sc SSF}&0.78&75.7&69.0&92.6&\bf75.1&\bf99.4&\bf91.8&\bf90.2&52.9&87.4&95.9&\bf87.4&75.5&75.9&62.3&\bf53.3&80.6&77.3&54.9&29.5&37.9\\
{\sc Adapter-P}&0.56&75.5&73.2&90.1&69.6&99.2&91.1&84.9&56.0&86.6&94.8&82.5&75.8&82.9&63.9&49.7&79.7&81.7&55.5&31.6&42.2 \\
{\sc AdaptFormer}&0.56&76.7&73.8&92.3&72.7&99.3&91.6&89.1&56.5&87.8&95.5&84.9&75.2&83.3&62.5&52.4&\bf81.7&86.2&\bf55.9&34.4&40.2\\
{\sc BitFit}&0.39&65.2&72.8&87.0&59.2&97.5&85.3&59.9&51.4&78.7&91.6&72.9&69.8&61.5&55.6&32.4&55.9&66.6&40.0&15.7&25.1\\
{\sc FacT-TT}&0.30&76.7&73.4&91.0&72.4&99.2&91.4&90.1&\bf56.6&87.3&94.7&84.5&75.8&83.0&64.9&51.3&81.4&\bf87.4&53.2&33.5&44.3\\
{\sc VPT-Shallow}&0.24&67.8&77.7&86.9&62.6&97.5&87.3&74.5&51.2&78.2&92.0&75.6&72.9&50.5&58.6&40.5&67.1&68.7&36.1&20.2&34.1\\
{\sc Compacter}&0.15&74.2&71.9&89.0&69.7&99.1&90.7&82.7&56.1&86.0&93.5&82.4&75.3&80.2&63.4&47.4&77.2&78.1&53.5&27.3&39.8\\
{\sc Bi-LoRA}&0.14&76.7&72.1&91.7&71.2&99.1&91.4&\bf90.2&55.8&87.0&95.4&85.5&75.5&83.1&64.1&52.2&81.3&86.4&53.5&36.7&\bf44.4\\
{\sc Bi-AdaptFormer}&0.071&\bf77.0&74.1&92.4&72.1&99.3&91.6&89.0&56.3&\bf88.2&95.2&86.0&\bf76.2&\bf83.9&63.6&53.0&81.4&86.2&54.8&35.2&41.3\\
\hline
\specialrule{0em}{1pt}{1pt}
\multicolumn{22}{l}{\emph{Pear}}\\
\hline
\specialrule{0em}{1pt}{1pt} {\sc Pear-LoRA} &0.07&76.7&72.8&92.0&71.3&99.1&91.3&89.4&55.9&86.9&95.5&84.5&75.5&83.0&64.2&51.8&\bf82.1&86.0&52.8&\bf38.2&43.9\\
\specialrule{0em}{1pt}{1pt} {\sc Pear-AdaptFormer} &\bf0.035&\bf77.0&74.3&\bf92.9&71.6&\bf99.4&91.3&89.6&56.4&\bf88.2&95.1&85.5&\bf76.3&\bf84.1&64.7&52.2&81.6&86.9&54.7&34.5&41.3\\
\bottomrule
\end{tabular}
}
\end{center}
\caption{\textbf{Results on the VTAB-1K benchmark}. "Avg. Acc." signifies the mean performance across three categories. "Size" indicates the average number of trainable parameters within the backbones for each task, excluding the classification heads which account for 0.14 MB per task on average. $^\dag$ indicates results sourced from~\citep{zhang2024NOAH} where inputs are standardized.}
\label{vtab}
\vspace{0pt}
\end{table*}

\section{Experiments}

\subsection{VTAB-1K Benchmark}

We assessed Pear on the VTAB-1K benchmark \citep{zhai2019vtab}, which evaluates general visual representations across 19 diverse tasks grouped into Natural, Specialized, and Structured categories. Each task uses 1000 examples, and we report top-1 accuracy on test sets. We compare our method with PEFT methods including VPT \citep{jia2022VPT}, NOAH \citep{zhang2024NOAH}, AdaptFormer \citep{chen2022adaptformer}, BiTFit \citep{zaken2021bitfit}, FacT-TT \citep{jie2023fact}, FourierFT \citep{gao2024fourierft}, MoRA \citep{jiang2024MoRA}, LoRA \citep{hu2021lora}, Bi-LoRA \citep{jie2023revisiting} and Bi-AdaptFormer \citep{jie2023revisiting}. For AdaptFormer and LoRA, the hidden dimension $r$ is fixed at 8. For FacT-TT, the rank $r$ is searched within the range of $\{8, 16, 32\}$. Both FourierFT and MoRA are implemented with parameter counts and training hyper-parameters aligned to those of LoRA. Following \citep{jie2023revisiting} and \citep{zhang2024NOAH}, we employed AdamW as the optimizer, utilizing a learning rate of 1e-3 and a batch size of 64, and conducted training over 100 epochs. For Pear, we apply it to both LoRA and AdaptFormer due to its method-agnostic nature. The prune ratio for Pear is fixed at $50\%$ (i.e., half of the adapters are pruned and then shared). Result for a task is searched in three scenarios: plain Pear, Pear with DA, and Pear with SBA. In SBA, we manually set both $C_1$ and $C_2$ as $0.5$. We also utilize the quantized adapters proposed by Bi-LoRA with a hidden dimension $d=32$. For identifying redundant parameters, we choose the first-order Taylor expansion for its efficiency and strong approximation capability \citep{molchanov2016pruningForResourseEfficientInference} \citep{kwon2022fast_taylor_prune_for_transformers}. We use 10 epochs as warm-up epochs to obtain the weights and gradients of the adapters, followed by fine-tuning with Pear over the subsequent 90 epochs. All methods employ the ViT-B/16 pre-trained on ImageNet 21K, following \citep{jie2023revisiting} and \citep{jie2023fact}.

The results are presented in Table \ref{vtab}, from which we can conclude that: Firstly, Pear boasts the smallest adapter size measured in MB, indicating a remarkable storage efficiency compared to other methods. Secondly, Pear achieves state-of-the-art performance despite utilizing the minimal parameter size, showcasing Pear's ability to maintain performance after heavily pruning. Overall, Pear achieves extremely high performance with a very small number of parameters, comparable to the state of the art. Therefore, Pear can deliver adaptation effects several times its own parameter count when faced with limited computational resources (restricted GPU memory or lighter devices), which makes it possible to run larger models on smaller devices.

\begin{table}[htb]
\centering
\resizebox{0.48\textwidth}{!}{
\begin{tabular}{l *{4}{c}} 
\toprule
Method & Average & Natural & Specialized & Structured \\\hline
\specialrule{0em}{1pt}{1pt}
Full & 74.99 & 79.2 & 86.2 & 59.7 \\
Linear & 62.60 & 73.5 & 80.8 & 33.5 \\
VPT-Deep & 71.55 & 76.8 & 84.5 & 53.4 \\
VP-LoRA & 75.97 & 81.3 & 84.9 & 61.7 \\
VP-AdaptFormer & 76.39 & 81.9 & 86.1 & 61.3 \\
\specialrule{0em}{1pt}{1pt}
\hline
\specialrule{0em}{1pt}{1pt}
Pear-LoRA & 76.68 & 81.7  & 85.6 & \textbf{62.7}\\ 
Pear-AdaptFormer & \textbf{76.98} & \textbf{82.2} & \textbf{86.3} & 62.5 \\\bottomrule
\end{tabular}
}
\vspace{3pt}
\caption{Comparison of the results on VTAB-1K between Pear and vanilla pruning. Both methods employ a pruning ratio of $ratio=0.5$, therefore half of the adapters are being pruned.}
\label{compare_to_vanilla_prune}
\end{table}

\subsection{Comparison to Vanilla Pruning}

As structural pruning, both vanilla pruning and Pear employ the same method to identify redundant parameters and prune those with low contributions according to a specified ratio. However, applying vanilla pruning to PEFT adapters removes the adapters at the corresponding positions, which indicates that these positions do not receive any adaptation after pruning. In fact, even with retraining, vanilla prune often results in performance that is still lower than without pruning. On the other hand, Pear addresses the issue of lack of continual adaptation by innovatively proposing a \textit{prune and share} strategy, as evidenced by the results on the VTAB in Table \ref{compare_to_vanilla_prune}.

We refer to vanilla pruning (VP) applied to LoRA and AdaptFormer as VP-LoRA and VP-AdaptFormer, respectively. Employing the same setting as Pear with a pruning ratio of $ratio=0.5$ and utilizing quantized adapters, ensuring a fair comparison. The results show that Pear significantly enhances the performance post-pruning, despite using the same amount of parameters as VP. This is due to Pear's innovative proposal of the \textit{prune-and-share} strategy, which allows for continual adaptation at the locations of pruned adapters. Additionally, leveraging the knowledge checkpoint technique we proposed under the Pear framework, we can still utilize the information contained in previously pruned adapters. This enables Pear to achieve much higher performance than VP with the same number of retrain epochs (an improvement of $0.6\%$ to $0.7\%$). Overall, this demonstrates that Pear achieves extreme storage efficiency and great performance, pushing the parameter-efficiency of current PEFT methods to its limit.

\begin{figure}
    \centering
    \includegraphics[width=0.8\linewidth]{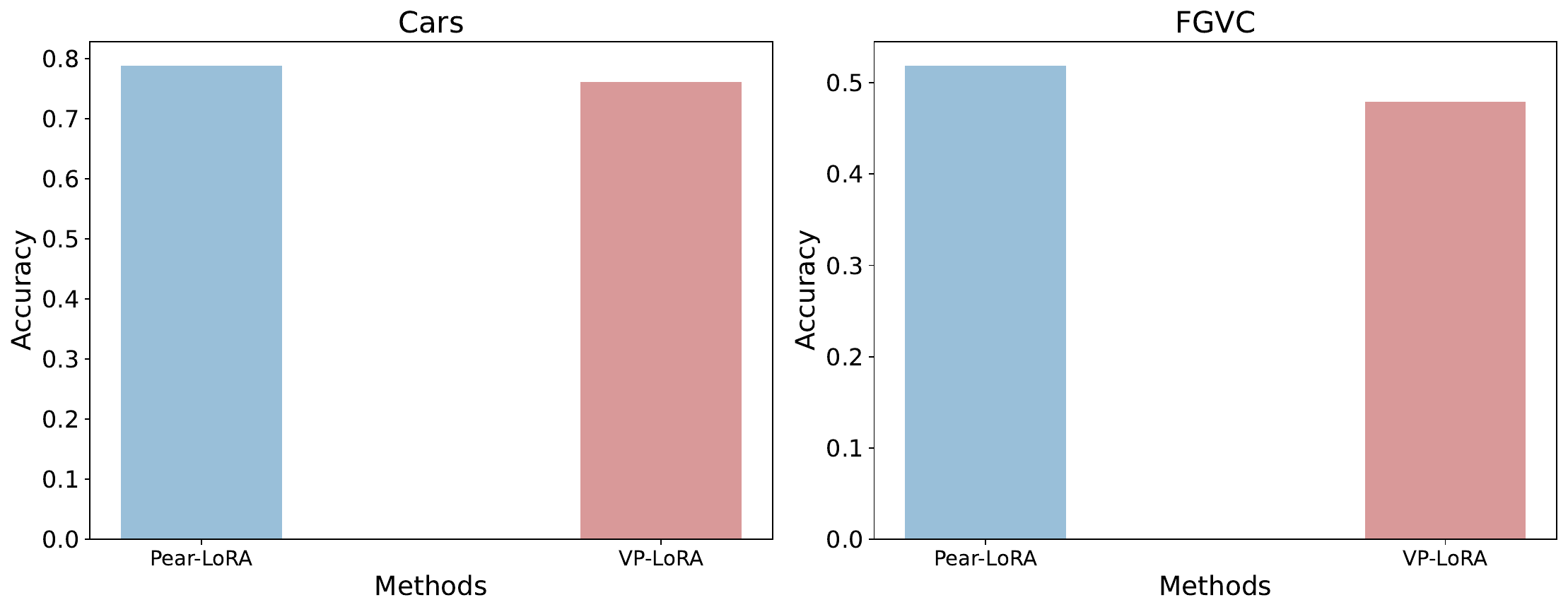}
    \caption{Comparison between Pear and vanilla pruning on FGVC and StanfordCars datasets. Vanilla pruning is denoted as VP here.}
    \label{cmp}
\end{figure}

Additionally, we conduct two more experiments on datasets out of VTAB-1K benchmark in Fig. \ref{cmp} to further provide an understanding of the advantage of Pear. The adapters used here are not quantized.
\begin{table}[htb]
\centering
\resizebox{0.48\textwidth}{!}{
\begin{tabular}{l *{4}{c}} 
\toprule
Method & Average & Natural & Specialized & Structured \\
\hline
\specialrule{0em}{1pt}{1pt}
LoRA: & & & & \\
\quad Pear & 76.28 & 81.4 & 85.3 & 62.1 \\
\quad Pear + DA & 76.50 & 81.6 & 85.4 & 62.5 \\
\quad Pear + SBA & 76.52 & 81.4 & 85.5 & 62.6 \\
\hline
\specialrule{0em}{1pt}{1pt}
AdaptFormer: & & & & \\
\quad Pear & 76.81 & 82.1 & 86.1 & 62.3 \\
\quad Pear + DA & 76.76 & 82.2 & 86.1 & 62.0 \\
\quad Pear + SBA & 76.81 & 82.0 & 86.2 & 62.2 \\
\bottomrule
\end{tabular}
}
\vspace{3pt}
\caption{Ablation of the search space of Pear. It can be observed that utilizing knowledge checkpoint generally improves the model's performance in certain tasks, thus showing its advantage of improving performances without any extra parameters.}
\label{ablation}
\vspace{-20pt}
\end{table}

\subsection{Ablation}

We provide the separate results of Pear in the search space in Table \ref{ablation}. We can observe that the performance of Plain Pear (i.e., only prune and share without knowledge checkpoint) is already significantly better than that of vanilla pruning. However, by employing the knowledge checkpoint, we can retain the information from the pruned adapters, allowing for more comprehensive adaptation on certain tasks and, consequently, higher performance.

\section{Conclusion}

In this paper, we propose Prune and Share (Pear), a novel pruning methods that enables continual adaption to the whole model even after pruning half of the parameters. Pear achieves a superior performance compared to vanilla pruning while utilizing the same amount of parameters. Additionally, we propose a strategy termed as knowledge checkpoint to preserve the information of the pruned parameters, which further boost the performance without any parameter overhead. We validate the effectiveness of Pear by conducting experiments on tasks from visual adaptation benchmark, showcasing its visual adaption capability.

\clearpage
\bibliography{ref}
\end{document}